\documentclass[twocolumn,10pt]{asme2e}

\usepackage{graphicx}
\usepackage{subfigure}
\usepackage{amsmath, amssymb}
\usepackage[table]{xcolor}

\confshortname{WINVR2010}
\conffullname{ASME 2010 World Conference on Innovative Virtual Reality}

\confdate{12-14}
\confmonth{May}
\confyear{2010}
\confcity{Ames, Iowa}
\confcountry{USA}

\papernum{WINVR2010-3729}

\title{Integrating digital human modeling into virtual environment for ergonomic oriented design}

\author{{\tensfb Liang MA}\thanks{Address all correspondence to this author.}, {\tensfb Damien CHABLAT}, {\tensfb Fouad BENNIS} 
    \affiliation{
	IRCCyN CNRS UMR 6597\\
	Ecole centrale de Nantes\\
	1 Rue de la No\"{e}\\
	44321 Nantes Cedex 3, France\\
    Email: \{liang.ma,damien.chablat,fouad.bennis\}@irccyn.ec-nantes.fr
    }	
}

\author{    
    {\tensfb Bo HU}, {\tensfb Wei ZHANG}      
    \affiliation{
	Department of Industrial Engineering\\
	Tsinghua University\\
	100084, Beijing, CHINA\\
	Email: hu-05@tsinghua.edu.cn\\
	zhangwei@tsinghua.edu.cn
    }
}

\begin{document}

\maketitle    

%%%%%%%%%%%%%%%%%%%%%%%%%%%%%%%%%%%%%%%%%%%%%%%%%%%%%%%%%%%%%%%%%%%%%%
\begin{abstract}
{\it Virtual human simulation integrated into virtual reality applications is mainly used for virtual representation of the user in virtual environment or for interactions between the user and the virtual avatar for cognitive tasks. In this paper, in order to prevent musculoskeletal disorders, the integration of virtual human simulation and VR application is presented to facilitate physical ergonomic evaluation, especially for physical fatigue evaluation of a given population. Immersive working environments are created to avoid expensive physical mock-up in conventional evaluation methods. Peripheral motion capture systems are used to capture natural movements and then to simulate the physical operations in virtual human simulation. Physical aspects of human's movement are then analyzed to determine the effort level of each key joint using inverse kinematics. The physical fatigue level of each joint is further analyzed by integrating a fatigue and recovery model on the basis of physical task parameters. All the process has been realized based on VRHIT platform and a case study is presented to demonstrate the function of the physical fatigue for a given population and its usefulness for worker selection.}
\end{abstract}

\section{Introduction}
Although automation techniques have been used more and more prevalent in industrialized countries, there are still large requirements of manual handling operations, especially for assembly and maintenance operations. For this reason, ergonomic oriented design should be carried out to take account of human capacities and limitations as early as possible. Therefore, work design and evaluation is very important to avoid potential musculoskeletal disorders (MSDs) and to determine appropriate work-rest schedule. 

In order to accelerate the ergonomic evaluation procedure, digital human modeling (DHM) techniques have been developed and integrated into computer aided design systems. In this way, work evaluation can be carried out for the overall population using anthropometric database; the high cost of physical mock-up can be avoided; the evaluation efficiency is much more improved in comparison to conventional evaluation approaches in workplace; potential ergonomic issues can be decreased. There are several commercial tools available in the market, for example, 3DSSPP\cite{Chaffin1999}, Jack\cite{BADLER1999}, AnyBody\cite{damsgaard2006ams}, Santos\texttrademark\cite{yang2007vpp}, etc. In these DHM tools, a simulated human is often used as a geometric representation, visual appearance of the real human being in computer graphical simulation \cite{granieri1995shv}, and furthermore some ergonomic and biomechanical analysis can be carried out using those tools\cite{witmer1998measuring,honglun2007rvh}.  However, as pointed out by \cite{chryssolouris2000vrb}, this approach cannot provide a mimic motion due to its robotic background. Sometimes, it requires to place the manikin in the simulation environment and to define all the motions manually, which is time consuming as well.

With the development of computer graphic techniques and human computer interaction techniques, virtual reality (VR) techniques have been used widely in different application fields, such as way finding and navigation techniques, object selection and manipulation, education, training, etc. Integrating visual, auditory and haptic systems, they are used to enable interactions between virtual world and the user \cite{stanney2003usability, loomis1999immersive, magnenatthalmann2006ivh}. Since 1990s, VR has been used in product design and manufacturing to facilitate the manufacturing procedure simulation, from the very initial conceptual design via operations management to final manufacturing processes \cite{shukla1996virtual, mujber2004vra}. Meanwhile, VR technique can also be used crossing different levels of an enterprise, form physical operations (e.g. virtual assembly\cite{jayaram1997vau,jayaram2007ics}) to human resource planning \cite{duffy2000concurrent}. Due to its immersive simulation and advanced interaction peripherals, it is believed that VR system can be used for ergonomic analysis\cite{whitman2004vri}. However, there are still some limitations for a VR system. 
Firstly, to create an immersive virtual environment with a high level of presence is very expensive. Also, it is often very difficult to obtain force interactions between the user and the virtual environment (VE) with high fidelity. Secondly, even using a VR system with high level of presence, it is still not easy to carry out ergonomic analysis for a given population, since different subjects might have different anthropometric parameters and different biomechanical parameters. With the engagement of only several subjects in VR applications, it cannot reveal all the design flaws.

Integrating virtual human techniques into VR applications can benefit the advantages from both sides \cite{zeltzer1992hfm}. Several studies have been done in the field of ergonomics\cite{wilson1999vea}, biomechanical analysis\cite{VSR2004}, and other aspects\cite{gutierrez2004mai}. Till now,  most of the trials focus on the visual representation of virtual human in VR\cite{ieronutti2007employing}, where virtual human is used to increase the presence level of the VR system. In our study, we are trying to use this approach for physical fatigue analysis of static manual handling operations, since it is believed that physical fatigue might be a very important reason resulting in MSDs. In this paper, the detailed research effort in combining virtual reality techniques and digital human simulation tools is described. In Section \ref{sec:vr}, a virtual reality platform for all the research activities is introduced, including its hardware structure and its peripheral systems; Section \ref{sec:dhm} describes briefly the modeling procedure of a virtual human and its work flow for fatigue analysis; Section \ref{sec:case} refers to an application case in EADS.

%Integration of fatigue and recovery model
\section{Virtual Reality Platform} \label{sec:vr}
\begin{figure}[htbp]
	\centering
		\includegraphics[width=0.450\textwidth]{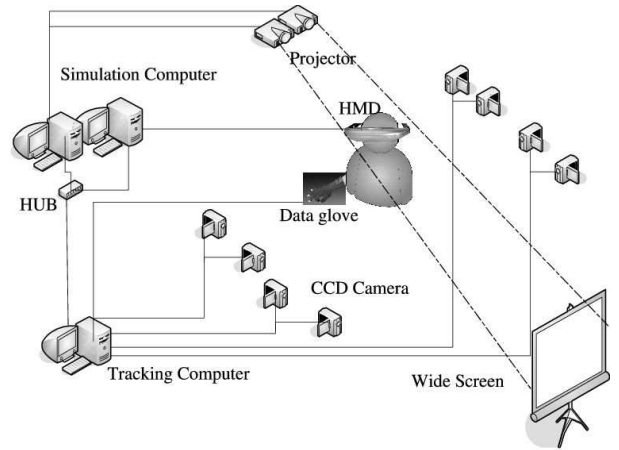}
	\caption{Illustration of the hardware system for the VRHIT platform}
	\label{fig:hardware}
\end{figure}

The VR platform is established in the laboratory of Virtual Reality \& Human Interface Technology (VRHIT) in Tsinghua University. The platform is developed for doing research on work design in virtual working environment. The overall system structure is illustrated in Figure \ref{fig:hardware}.

%Simulation
The virtual simulation system is equipped with simulation computers, a wide screen projection, and 5DT HMD 800-26. MultiGen Creator is used to create virtual objects, while Vega is to employ MultiGen Creator to convert a CAD model into a virtual reality model \cite{ren2004modeling}. The virtual simulation system is able to load VR models and display via either the projection system or HMD. Stereo display is possible for some desktop applications.

\begin{figure}[htbp]
	\centering
		\includegraphics[width=0.450\textwidth]{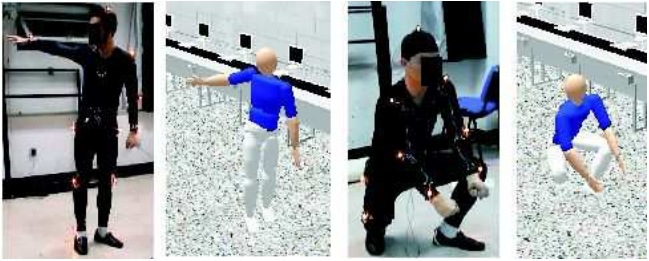}
	\caption{Self-developped motion capture system and its simulation system}
	\label{fig:motionsincapture}
\end{figure}

Different peripherals attached to this system include motion tracking systems and data gloves. Two magnetic tracking systems (Fastrak$^{\circledR}$ and Potroit$^{\circledR}$ of Polhemus co.) and a self-developed optical motion tracking system are available for this platform. A participant wearing the motion capture suit and performing certain physical tasks is shown in Figure \ref{fig:motionsincapture}. The optical capture
system is equipped with eight CCD cameras around the work space. In our case, some operations with high physical demands are under consideration, and in these operations, only static operations or slow movements are involved or considered. The overall capture system works at the frequency 25 $Hz$, which satisfies the minimum requirement to provide sufficient update rate of the simulation image, especially for quasi-static postures, and it provides sufficient detailed analysis of the static human body motion. Two data gloves, a 5DT Data Glove 5 Ultra and a self-developed data glove with haptic feedback, are in our possession to provide detailed gesture of the hands and feedback information while manipulating virtual objects.

\begin{figure}[htp]
\centering
\subfigure[subject]
	{\label{fig:subject}
	 \centering
	 \includegraphics[width=0.18\textwidth]{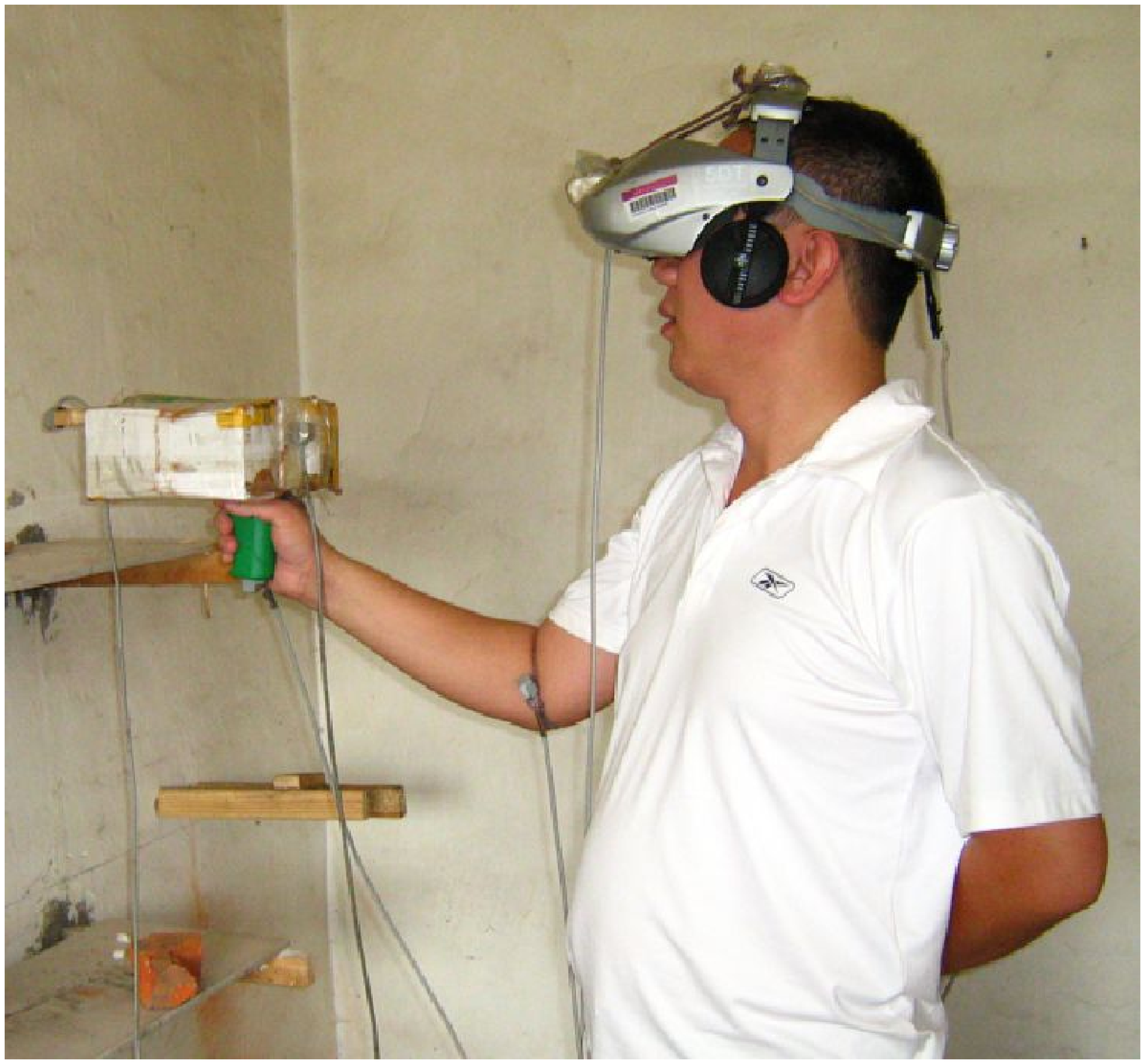}
	}\qquad
\subfigure[Virtual environment]{\label{fig:ve}\includegraphics[width=0.18\textwidth]{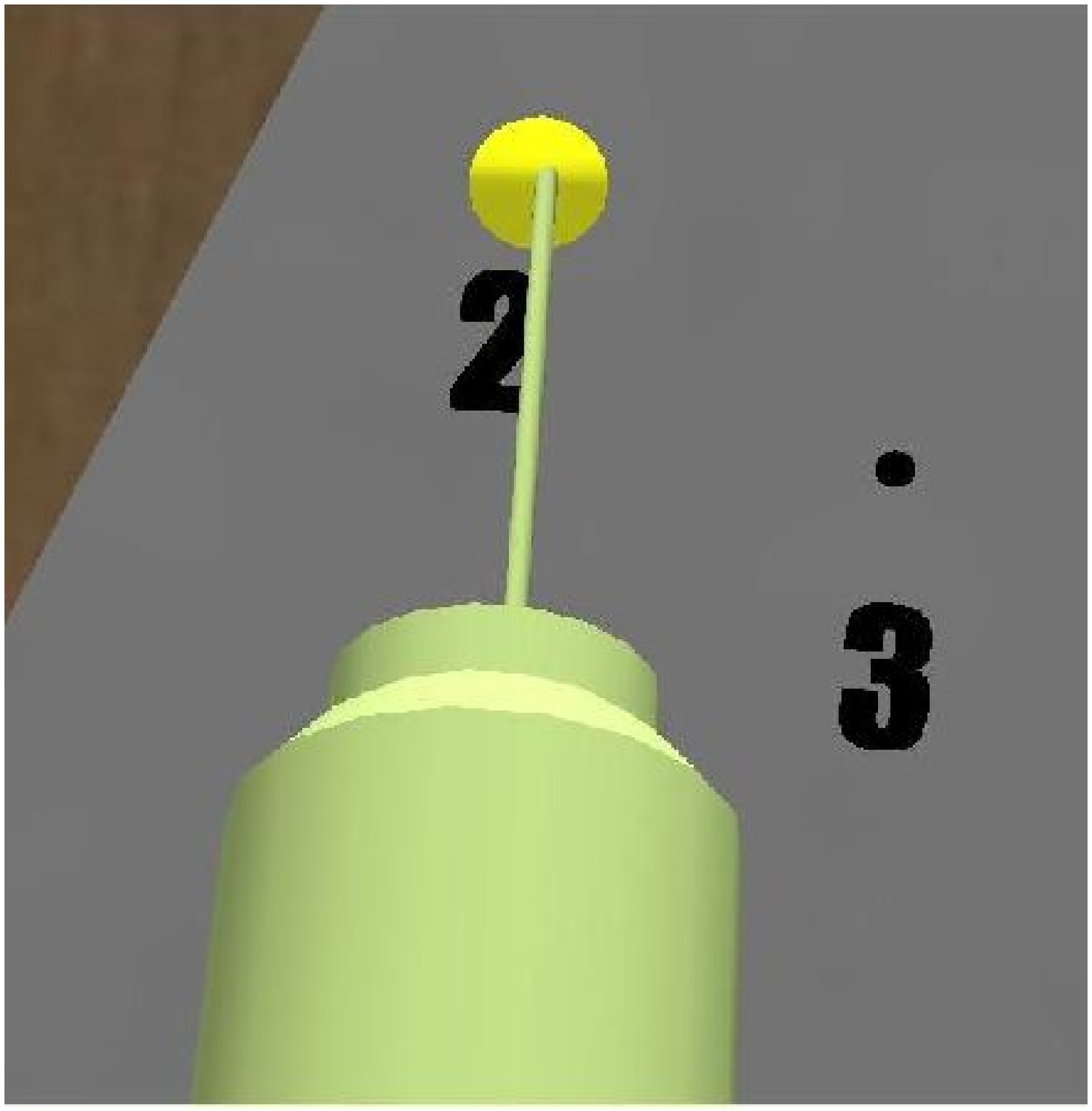}}\\
\caption{Subject wearing HMD in virtual environment}
\end{figure}

%Motion capture
The objective of the VRHIT experiment platform is to enable interactions between human and the simulated manufacturing environment and to realize interactive industrial work design taking consideration of human factors and ergonomics. This platform has been used in \cite{WANG2006}, \cite{hu2008development}, and \cite{ma2009framework}, for verifying a control panel design, assembly work design in heavy industry, and work efficiency evaluation, respectively. In this paper, it is mainly used to capture the static posture of a simulated drilling operation (see Figure \ref{fig:subject}). The posture is then analyzed by inverse dynamics based on digital human modeling techniques.

\section{Digital human modeling} \label{sec:dhm}
\begin{figure}[htbp]
	\centering
		\includegraphics[width=0.45\textwidth]{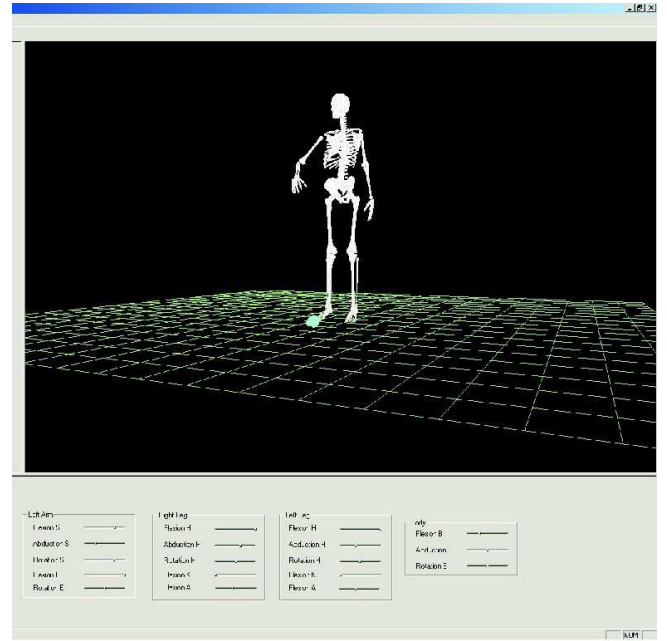}
	\caption{Geometrical modeling and graphical modeling of the virtual human}
	\label{fig:Interface}
\end{figure}

In this study, the human body is kinematically composed of a series of revolute joints. The Modified Denavit-Hartenberg (modified DH) notation system \cite{khalil02} is used to describe the movement flexibility of each joint and the linkage relation among the joints. According to the joint function, one natural joint can be decomposed into one, two, or three revolute joints. Each revolute joint has its rotational joint coordinate, labeled as $q_i$, with joint limits: the upper limit  $q_i^U$ and the lower limit $q_i^L$. A general coordinate $\mathbf{q}=[q_1,q_2,\ldots,q_n]$ is defined to represent the kinematic chain of the skeleton. Overall, a virtual human is geometrically modeled by $n=28$ (3 in shoulder, 2 in elbow, 3 in thigh, 1 in knee, 1 in ankle, 2 in waist, 4 in spine, 2 in neck) revolute joints to reproduce the mobility of all the key joints on the human body. The skeleton system (in Figure \ref{fig:Interface}) of the virtual human is graphically modeled using OpenGL and controlled by direct kinematics, inverse kinematics, or loading motion capture data.

The dimension and dynamic parameters of each body part can be obtained or calculated from anthropometry database, and meanwhile biomechanical strengths of each joint can be obtained from the literature in Occupational Biomechanics \cite{Chaffin1999}. In this way, the virtual human is modeled as a multi-rigid-body system, and the forces and torques at each key joint can be determined using inverse dynamics.

\begin{figure}[htbp]
	\centering
		\includegraphics[width=0.450\textwidth]{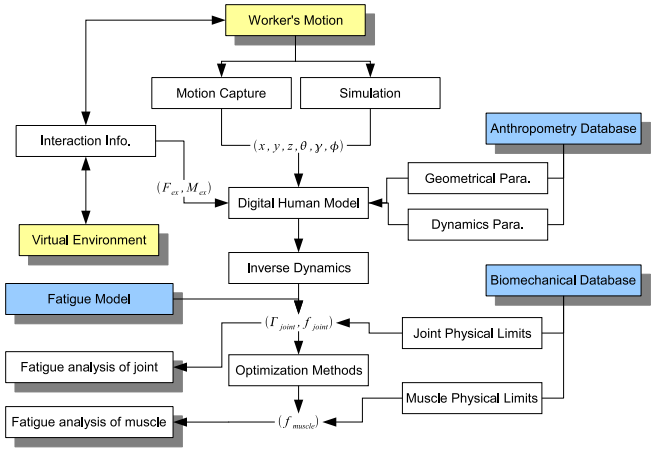}
	\caption{Work flow for biomechanical analysis}
	\label{fig:workflow}
\end{figure}

The work flow for physical fatigue analysis of manual handling tasks is shown in Figure \ref{fig:workflow}. In order to carry out work analysis, it is necessary to collect human motion and interaction between human and the environment. In our case, it is possible to obtain human motion from motion capture system, and then the motion information is transferred in the form of general coordinates $\mathbf{q}$. Human motion can
also be obtained via human simulation tools. Large modeling efforts have been done in IOWA Santos and HUMOSIM \cite{zhou2009simulating} to predict realistic postures and motions. Based on the techniques used in these two projects, a multi-objective posture prediction method is used to predict realistic postures with consideration of fatigue in \cite{ma2009multi}. 
Interaction information is obtained either via haptic interface or via computer simulation. In some cases, especially for cases with high physical demand, the VR system is not able to provide precise and accurate force feedback to the user, and the external load is estimated or provided by real objects. In this way, inverse dynamics can be carried out to determine the force and the torque at each joint.  After determining the force and the torque load at each joint, it is possible to carry out joint fatigue analysis. If it goes further, it might be possible to determine the force of each engaged individual muscle as well.

Physical fatigue is defined as ``any exercise-induced reduction in the capacity to generate force or power output'' \cite{vollestad1997mhm}. After knowing the physical load at each joint and the joint's strength, the muscle fatigue model proposed in \cite{ma2008nsd} is used to assess the physical fatigue during manual handling operations. This model can be described by a differential equation in Eq.\ref{eq:MomentDiff}.
\begin{equation}
\label{eq:MomentDiff}
			\frac{d\Gamma_{cem}(t)}{dt} = -k \frac{\Gamma_{cem}(t)}{\Gamma_{max}}\Gamma_{load}(t)
\end{equation}
Where $\Gamma_{max}$ represents the maximum joint moment capacity; $\Gamma_{cem}$ represents the current maximum joint moment capacity during a static exertion; $\Gamma_{load}$ is static joint moment load; $k$ represents the joint specified fatigue rate and it could vary across individuals and joints. $m=1/k$ indicates the fatigue resistance of a given joint. Using this model, the external task related parameters can be taken into consideration to assess the fatigue of a joint during a given manual handling operation. 

Meanwhile, a muscle recovery model using the same parameters is used to describe the recovery process of the joint strength during rest period. It is mathematically described using Eq. \ref{eq:FcemDiffRec}, where $R$ is the recovery rate. This model is extended from a recovery model in \cite{wood1997mfd}. 
\begin{equation}
\label{eq:FcemDiffRec}
			\frac{d\Gamma_{cem}(t)}{dt} = R (\Gamma_{max}-\Gamma_{cem}(t))
\end{equation}

It should be mentioned that $k$ and $R$ are parameters to describe the fatigue properties of an individual or a population. They have almost the same function as the height and weight. However, they are very difficult to be measured directly. For the fatigue resistance, it has been mathematically analyzed in \cite{ma2008nsd, ma2009contribution}. For shoulder joint, it could be regressed from several existing maximum endurance
time models (MET), and it has a theoretical value 0.7562 (SD=0.4347). 

In order to carry out the fatigue evaluation around a joint, it is necessary to obtain the necessary strength data and endurance data for a population. Currently, there are several biomechanical models available after the measurement of maximum joint strength under different conditions in the literature. However, these models are not sufficient enough to construct a complete database for a given population, since there are
lots of influencing factors determining the joint strength. Thus, more effort should be contributed to gathering all necessary data. Concerning the fatigue resistance and recovery rate, experimental validation is still under running to complete the validation of fatigue/recovery model.

For each joint, after a manual handling operation, the fatigue could be assessed using a fatigue index which indicates the change of the maximum joint moment strength (Eq. \ref{eq:index}).

\begin{equation}
\label{eq:index}
		fatigue= \dfrac{\Gamma_{max}-\Gamma_{cem}}{\Gamma_{max}}
\end{equation}

\section{Application} \label{sec:case}
\subsection{Drilling case in EADS}
\begin{figure}[htbp]
	\centering
		\includegraphics[width=0.450\textwidth]{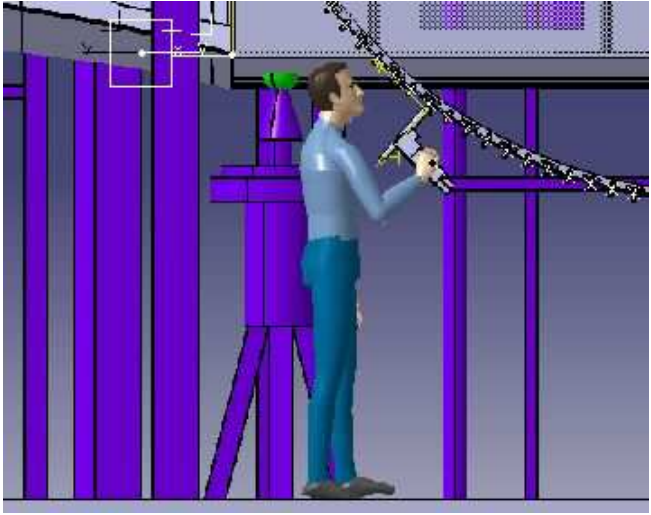}
	\caption{Drilling operations in the case study}
	\label{fig:Work}
\end{figure}

A drilling case in European Aeronautic Defence and Space (EADS) is taken for the case study. The drilling operation is graphically shown in Figure \ref{fig:Work}. The task is to assemble two fuselage sections with rivets. One part of the job consists of drilling holes all around the aircraft cross section. The number of the holes could be up to 2,000 on an orbital fuselage junction of an airplane. The drilling machine has a weight around 5 $kg$. The drilling force applied to the drilling machine is around 49 $N$. In general, it takes 30 seconds to drill a hole.  There are several ergonomic issues due to the high physical demands and vibration in drilling operations, and the physical fatigue caused by the physical operation might be one limitation for the work schedule design. 

This drilling task is simplified to two study case. One case is to assess the fatigue of a continuous drilling task with duration 180 seconds.  The other case is to select suitable workers for a drilling task with five work cycles (60 second drilling and 30 seconds rest). 
\subsection{Case 1: fatigue assessment}
The first case at first carried out using the virtual reality platform.  A total of 40 male industrial workers participated in the experiment. They were asked to maintain a simulated drilling operation for 180 seconds. During this period, the subjects were asked to keep the static posture as well as possible, while the posture was captured by magnetic tracking devices and mapped to virtual human for further theoretical fatigue analysis. 

During the drilling operation, the load of the drilling machine was provided by a simulated drilling tool, which was made of concrete with a weight around 2.5 $kg$ to avoid magnetic distortion caused by ferrous material. Maximum output strengths were measured by a force measurement device before and right after the simulated drilling operations. Physical fatigue was characterized by the reduction of the joint strength along time relative to the initial maximum joint strength (see Eq. \ref{eq:index}). It has been found that three out of the 40 subjects could not sustain the external load for 180 seconds, and 34 subjects had a shoulder joint fatigue resistance (Mean=1.32, SD=0.62). The physical strength has been measured in simulated job static strengths, and the reduction in the operation varies from 32.0\% to 71.1\% (Mean=53.7\% and SD=9.1\%) \cite{ma2009contribution}. 

Meanwhile, the same drilling operation was also analyzed in virtual human simulation. In order to simplify the simulation process, the average posture (flexion angles of elbow and shoulder) among the 40 subjects was used to determine the shoulder joint strength according to Occupation Biomechanics\cite{Chaffin1999}. The fatigue resistance for calculating the fatigue theoretically was obtained by comparing the theoretical fatigue model with other models in the literature ($\bar{m}$=0.7562, $\sigma$=0.4347) \cite{ma2009contribution}. It is believed industrial workers have a higher fatigue resistance due to the training effect \cite{hawley1997training}. For this reason, the simulation was executed with fatigue resistances superior to the average fatigue resistance, and the reduction of the shoulder joint strength was given in Table \ref{tab:fatigue} (Mean=51.7\%, SD=12.1\%). It has been found that the simulated result was similar to the realistic measured fatigue, which means the theoretical approach in virtual human simulation might give an acceptable assessment for the specified drilling operation of a given population.

\begin{table}[htbp]
	\caption{Normalized torque strength reduction for the population with higher fatigue resistance. $S$ and $\sigma$: mean and SD of shoulder joint moment strength; $\bar{m}$ and $\sigma_m$: mean and SD of fatigue resistance}
	\centering
		\begin{tabular}{lccccc}
			\hline\noalign{\smallskip}			 $\dfrac{\Gamma_{max}-\Gamma_{180}}{\Gamma_{max}}$&$S-2\sigma$&$S-\sigma$&$S$&$S+\sigma$&$S+2\sigma$\\
			\hline\noalign{\smallskip}	
			$\bar{m}$&-&-&69.9\%&62.5\%&56.3\%\\
			$\bar{m}+\sigma_m$&-&63.2\%&53.2\%&46.4\%&40.8\%\\
			$\bar{m}+2\sigma_m$&64.9\%&51.9\%&43.0\%&36.7\%&31.9\%\\
			\hline\noalign{\smallskip}
		\end{tabular}
	\label{tab:fatigue}
\end{table}

In the simulation of this case, only different shoulder joint strengths and different shoulder fatigue resistances were taken into consideration. Some other influencing factors were neglected, for example, different body weights, different heights, and some others. In fact, although there are definitely other factors influencing the simulation result, joint strength and joint fatigue resistance are two most important parameters for fatigue analysis. The following subsection will show the effects of different factors.

\subsection{Case 2: worker selection}
The second case was studied to verify the possible usability of virtual human simulation for worker selection. Worker selection is often a practical problem during work design. In work design, the task has often been predefined, and there might be no possibility to modify the task parameters due to strict constraints. In this situation, designers have to choose suitable workers fitting the task. In ergonomics, generally, the anthropometric parameters (e.g., height, weight), motion range, and biomechanical parameters (strength) are often used to create some criteria to select the worker \cite{Chaffin1999}. For operations with high physical demands, four parameters (height, weight, strength, and fatigue resistance) might generate potential influences on the selection. 

A drilling case was simulated with the help of digital human simulation to demonstrate effects of those parameters on physical fatigue operations. The drilling operation is composed of five work cycles. In each cycle, the worker has to perform the drilling operations for 60 $s$ continuously, and then takes a rest for 30 $s$. The total working time is 450 $s$.  The selection criteria would be to find the workers who might be able to complete the given task. 

In the simulation, the variation among the population is taken into consideration, since different heights, weights, strengths, and fatigue rates might generate potential influences on the evaluation results. A virtual human was created using different anthropometric data (height, weight) to cover 95\% of the population, while the average value and the standard deviation of the strength and fatigue resistance were also taken into consideration. For each factor, there are three levels: average level, lower level, and higher level.  For height, weight and strength, the lower level and the higher level were $(\bar{x}-2\sigma)$ and $(\bar{x}+2\sigma)$, respectively. $(\bar{x}-\sigma)$ and $(\bar{x}+\sigma)$ were used to represent the lower level and the higher level of the shoulder fatigue resistance. Using these factors at three different levels, the worker could be categorized into 81 different subgroups, and each subgroup has its own properties accordingly.

The maximum endurance time was predicted using the fatigue model and recovery model mentioned before, and it was determined while the joint capacity falls down to the current torque load. There are totally 81 endurance results for the drilling task, and the endurance results of the shoulder joint are shown in Table \ref{tab:Endurance}. 

\begin{table}[htbp]
	\centering
	\caption{Endurance time evaluation of the drilling case. H: height; W: Weight; S: Joint moment strength}
	\label{tab:Endurance}
	\begin{tabular}{ccclll}
		\hline
		 \multicolumn{6}{c}{Endurance time of the shoulder joint [second]} \\
		\hline
		\multicolumn{3}{c}{}&\multicolumn{3}{c}{Fatigue resistance}\\
		\cline{4-6} H&W&S&Large&Average&Small\\
		\hline
		Low&Light&Weak&414&328&219\\
		&&Average&$>450$&$>450$&$>450$ \\
		&&Strong&$>450$&$>450$&$>450$\\
		&Average&Weak&408&322&241\\
		&&Average&$>450$&$>450$&416\\
		&&Strong&$>450$&$>450$&$>450$\\
		&Heavy&Weak&401&316&144\\
		&&Average&$>450$&$>450$&412\\
		&&Strong&$>450$&$>450$&$>450$\\
		\hline
		Average&Light&Weak&409&324&216\\
		&&Average&$>450$&$>450$&417\\
		&&Strong&$>450$&$>450$&$>450$\\
		&Average&Weak&403&319&147\\
		&&Average&$>450$&$>450$&414\\
		&&Strong&$>450$&$>450$&$>450$\\
		&Heavy&Weak&399&314&139\\
		&&Average&$>450$&$>450$&410\\
		&&Strong&$>450$&$>450$&$>450$\\
		\hline
		Tall&Light&Weak&404&320&215\\
		&&Average&$>450$&$>450$&414\\
		&&Strong&$>450$&$>450$&$>450$\\
		&Average&Weak&400&316&219\\
		&&Average&$>450$&$>450$&411\\
		&&Strong&$>450$&$>450$&$>450$\\
		&Heavy&Weak&396&313&219\\
		&&Average&$>450$&$>450$&408\\
		&&Strong&$>450$&$>450$&$>450$\\
		\hline			
		\end{tabular}
\end{table}
%Influence from height and weight
Height and weight, both factors could generate slight influences on the endurance time. It could be observed that smaller people have longer endurance time in comparison to taller ones with the same strength and the same fatigue resistance, since the moment arm might be shorter. The moment load can also be influenced by the weight. Heavier ones might generate larger moment around joint and then shorten the endurance time. However, it should be noticed that the influence on the endurance time caused by the height and weight are relatively slight, and it can neglected. Therefore, these two factors are not significant parameters to choose suitable workers for a purely physical task.

% influence from strength and fatigue resistance
Obviously, the endurance is greatly influenced by the joint strength and the fatigue resistance. Larger joint capacity results in much longer endurance time at the same fatigue resistance level. Clearly, larger fatigue resistance enables us to maintain the physical task for a longer time at the same strength level. This observation conforms to our daily experience. Furthermore, if a worker could be characterized using the fatigue resistance and the joint strength, that means it might be promising to locate the worker in the 81 categories.  In this way, suitable worker could be selected for a given physical task.

The validation of the second case is still under construction, and the method and the difficulties will be discussed below. For the validation, the fatigue resistance, the recovery rate, and the joint moment strength of a subject should be measured. However, it is difficult to determine the joint moment strength since it depends on several factors, such as posture, movement velocity, etc\cite{black2007tdc}. It would be even more difficult to measure the fatigue resistance and the recovery rate for an individual. Although a experimental method has been proposed and realized in \cite{ma2009contribution}, this method is time-consuming. Furthermore, it would very expensive to obtain the fatigue resistance distribution of a given population, since enormous measurements will be carried out. An much more efficient method has to be found before the validation. Regarding the recovery rate, a large dispersion has been found in existing experimental models for determining work-rest schedule \cite{elahrache2009cra}. How to characterize the recovery rate is a much more challenging task for us. Therefore, only a simulation task is provided for this case. All the related parameters were obtained from the literature or by theoretical analysis. However, it is believed that a guidance function could be provided.

\section{Conclusions and perspectives}
In this paper, an approach of integrating digital human simulation into virtual reality application was introduced. The virtual human is modeled from geometrical level to graphical level, to virtually represent a real human in the simulation system. It does not only provide a virtual representation, but also a tool for assessing different aspects of the task. In this study, inverse dynamics and the muscle fatigue and recovery model are integrated to predict the physical fatigue and endurance time at joint level.  A case study was given for assessing physical fatigue of a drilling operation.  Using virtual human simulation in VR system, it is promising to predict the fatigue of a given population under static cases in virtual environment. Furthermore, worker selection can also be accomplished in VR systems, if personal properties in physical fatigue could be well determined.

Assessing manual handling operation in virtual environment is very useful for work design. The research mentioned in study was limited by the biomechanical model of the virtual human, because the fatigue resistance and the recovery rate for a given population have not been measured. We have made some effort and we will continue to do the research in order to find an efficient method to determine the fatigue properties of an individual. Once fatigue and recovery for an individual can be easily characterized by the measured fatigue resistance and recovery rate, a biomechanical database might be established for a given population, or different population groups (e.g., industrial workers, nurses, etc.), and then physical assessment could be further performed for different physical tasks involving different population groups.

\bibliographystyle{asmems4}

%%%%%%%%%%%%%%%%%%%%%%%%%%%%%%%%%%%%%%%%%%%%%%%%%%%%%%%%%%%%%%%%%%%%%%
\begin{acknowledgment}
This research is supported by the EADS and by the R\'{e}gion des Pays de la Loire (France) in the context of collaboration between the Ecole Centrale de Nantes (Nantes, France) and Tsinghua University (Beijing, P.R.China). The lead author would also like to thank \'{E}cole Centrale de Nantes for the financial support of the post-doctorate studies.
\end{acknowledgment}

\bibliography{Bibliography}

\end{document}